\documentclass[runningheads]{llncs}
\usepackage{graphicx}
\usepackage{tikz}
\usepackage{comment}
\usepackage{amsmath,amssymb} % define this before the line numbering.
\usepackage{color}
\usepackage{booktabs}
\usepackage{tabularx}
\usepackage{threeparttable}
\usepackage{booktabs}
\usepackage{multirow} 
\usepackage{wrapfig}
\usepackage{hyperref}

\usepackage[accsupp]{axessibility}  % Improves PDF readability for those with disabilities.

\begin{document}
% \linenumbers
\pagestyle{headings}
\mainmatter
\def\ECCVSubNumber{3511}  % Insert your submission number here

\title{DynaST: Dynamic Sparse Transformer\\for Exemplar-Guided Image Generation} % Replace with your title

%\begin{comment}
%\titlerunning{ECCV-22 submission ID \ECCVSubNumber} 
%\authorrunning{ECCV-22 submission ID \ECCVSubNumber} 
%\author{Anonymous ECCV submission}
%\institute{Paper ID \ECCVSubNumber}
%\end{comment}
%******************

\titlerunning{DynaST: Dynamic Sparse Transformer}
\author{Songhua Liu \and
Jingwen Ye \and
Sucheng Ren \and
Xinchao Wang$^*$}
\authorrunning{Liu et al.}
\institute{National University of Singapore\\
\email{\{songhua.liu,suchengren\}@u.nus.edu}, \email{\{jingweny,xinchao\}@nus.edu.sg}}
%******************
\maketitle

\renewcommand{\thefootnote}{\fnsymbol{footnote}}
\footnotetext[1]{Corresponding author.}
\renewcommand{\thefootnote}{\arabic{footnote}}

\begin{abstract}
One key challenge of exemplar-guided image generation
lies in establishing fine-grained correspondences
between input and guided images.
Prior approaches, despite the promising results,
have relied on either estimating
\emph{dense attention to compute per-point matching},
which is limited to only coarse scales due to the quadratic memory cost, 
or \emph{fixing the number of correspondences}
to achieve linear complexity,
which lacks flexibility. 
In this paper, we propose a dynamic sparse attention based Transformer model,
termed 
\emph{Dyna}mic \emph{S}parse \emph{T}ransformer~(\emph{DynaST}),
to achieve fine-level matching 
with favorable efficiency.
The heart of our approach is a novel dynamic-attention unit,
dedicated to covering the variation on the optimal number of tokens 
one position should focus on. 
Specifically, DynaST
leverages the multi-layer nature of Transformer structure,
and performs the 
dynamic attention scheme 
in a cascaded manner to refine matching results
and synthesize visually-pleasing outputs. 
In addition, we
introduce a unified training objective
for DynaST,
making it a versatile reference-based image translation framework 
for both supervised and unsupervised scenarios. 
%In addition, we show that DynaST is adaptable to configurations of training objectives for both supervised and unsupervised scenarios, making it a versatile reference-based image translation framework. 
Extensive experiments on three applications, pose-guided person image generation, 
edge-based face synthesis, and undistorted image style transfer,
demonstrate that DynaST achieves superior performance in local details,
outperforming the state of the art while reducing the computational cost significantly. 
Our code is available \href{https://github.com/Huage001/DynaST}{here}.
\end{abstract}

\section{Introduction}
\label{sec:intro}

Semantic-based conditional image generation 
refers to synthesizing a photo-realistic 
image with aligned semantic information,
and finds its application across
a wide spectrum of scenarios
including label-to-scene~\cite{isola2017image,wang2018high,park2019semantic,zhu2020sean,tang2020local,wang2021image,esser2021taming,tan2021efficient}, sketch-to-photo~\cite{lee2020reference,gao2020sketchycoco,li2017example},
and landmark-to-face~\cite{song2018geometry,zakharov2019few,tang2019cycle}.
Exemplar-guided image generation,
as a mainstream approach along this route, 
provides users with the flexibility to specify 
an image as reference to control the appearance, 
style, or identity for the output image,
and has recently received 
wide attentions across
academic and industrial communities. 

\begin{figure}[t]
  \centering
  \includegraphics[width=\linewidth]{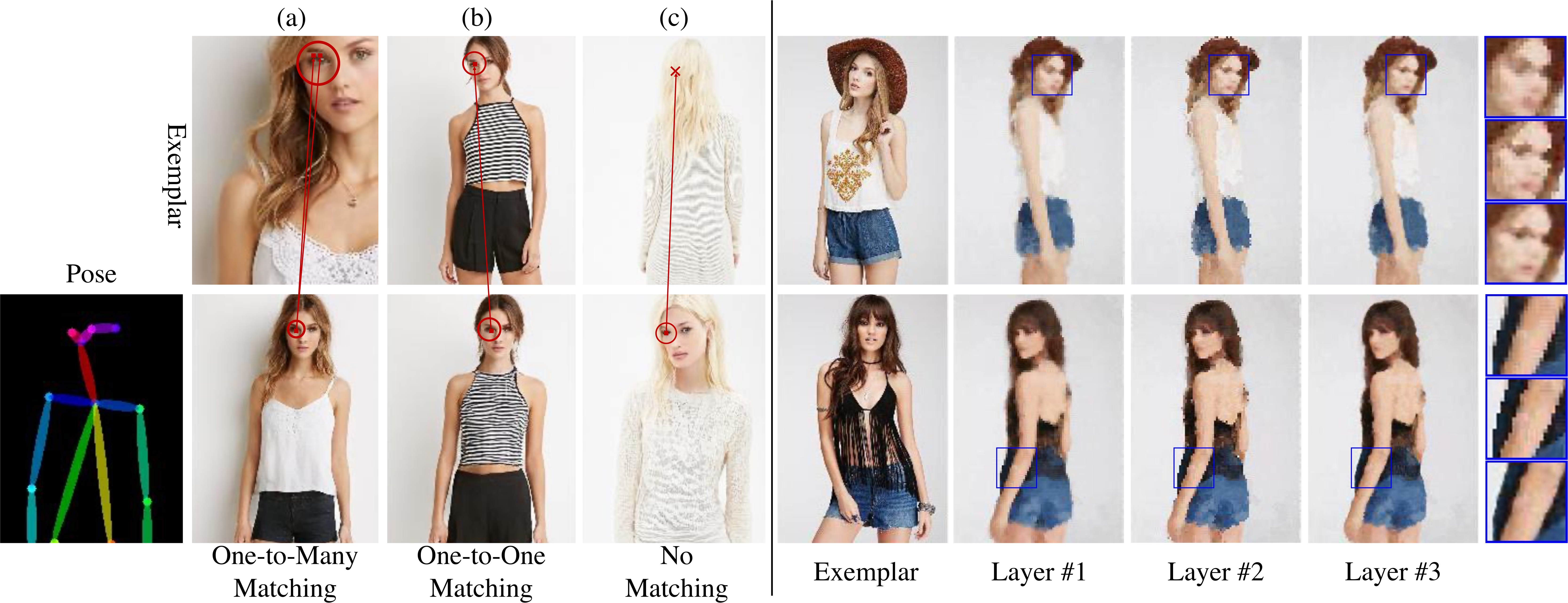}
%   \vspace{-0.8cm}
  \caption{Left: Different numbers of matching are required for 
  different query locations in 
  the exemplar-guided image generation task. 
  This fact has been largely overlooked by
  prior methods that impose 
  only static number of matching.
  The proposed DynaST, by contrast, is dedicated to handling
  such variations. 
  Right: Details in faces and edges can be refined with the propagation in the multi-layer structure of Transformer. Warping results of the exemplar images using attention maps in each layer are shown here. 
  }
%   \vspace{-0.8cm}
  \label{fig:teaser}
\end{figure}

The core problem of exemplar-guided image generation
lies in guiding input semantics
to focus on appropriate context in exemplars.
Early methods have largely relied on holistic convolution,
normalization, and non-linear transformation~\cite{ma2017pose,park2019semantic,wang2019example,albahar2019guided,zhu2019progressive,zheng2020example,tang2020xinggan,tan2021efficient}. 
Despite the promising results on global-style
migration and the favorable efficiency, 
these methods overlook the fine-grained local
details and thus lead to coarse results.
To account for local context in the 
exemplar-guided image generation, 
recent works~\cite{tangbipartite,zhang2020cross,zhan2021unbalanced} 
employ per-point attention mechanism to model spacial correspondence
between input and reference images. 
Nevertheless, hindered by the quadratic time and memory complexity, 
such dense matching operations, unfortunately,
again limit themselves to coarse scales, 
making them difficult to capture fine-grained details in reference images. 

To alleviate this issue, the works of~\cite{zhou2021cocosnet,zhan2021bi}
propose to fix the number of feature points
that each query position focuses on,
thereby achieving a linear-complexity model. 
The reason behind such a design
lies in that, 
each query position in target images is,
in reality, 
independent to and 
dissimilar to most points in the reference.
This rationale, in turn,
implies that
the matching between the target and reference
image is intrinsically sparse.

Unfortunately,  such the static number of 
correspondences, in many cases, 
fail to capture the dynamic nature of matching.  
As shown in Fig. \ref{fig:teaser},
different queries may end up having
different numbers of necessary matching:
in Fig.~\ref{fig:teaser} Left~(a) and (b), 
due to the scale variations, the highlighted query
location in the target image
corresponds to different numbers of locations in the reference; 
in Fig.~\ref{fig:teaser} Left~(c), however,
the query has no correspondence at all,
meaning that all points in the
exemplar would be negative samples. 

These facts motivate us to explore a more sophisticated technique to 
account for the dynamic matching inherent to the 
exemplar-guided image generation, 
while maintaining the sparsity 
to ensure computational efficiency.
%tackle the limitation on computational resources. 
To this end, we propose a novel Transformer-based model, termed \emph{Dyna}mic \emph{S}parse \emph{T}ransformer~(\emph{DynaST}). 
%The key idea in DynaST is to adopt a differentiable and learnable attention link pruning unit, to predict whether a matching candidate is 
%irrelevant, thereby encouraging precise matching results. 
%Specifically, the
%dynamic attention strategy further is
%backed up by a multi-layer structure as in~\cite{vaswani2017attention},
%that enables DynaST to evolve the
%sparse matching results in
%a cascade manner. 
The heart in DynaST is the dynamic sparse attention module in contrast to previous dense and static ones. 
Specifically, since the adopted attention strategy is sparse, a large number of potential matching candidates are dismissed. 
To alleviate this issue, we are inspired by the architecture of Transformer~\cite{vaswani2017attention} and back up the proposed attention strategy with a multi-layer structure, which enables DynaST to explore and evolve matching results in a cascade manner. 
As shown in Fig. \ref{fig:teaser} Right, we visualize the matching results by warping exemplar images using attention maps in each layer and observe that with the feature propagation in the multi-layer structure of Transformer, the model produces finer matching results, especially in local details such as faces and edges. 
To refine matching results progressively, each current matching candidate would pass through a differentiable and learnable attention link pruning unit in each layer, to predict whether it is an irrelevant correspondence. 

Consequently, (1) such dynamic pruning manner encourages more precise and cleaner matching results; 
(2) due to the effective higher-order dependency modeling capability of Transformer, DynaST is competent in aggregating relevant features and synthesizing high-quality outputs; and
(3) sparse attention in DynaST guarantees the efficiency of even full-resolution matching construction. 
Moreover, we introduce a unified training objective,
so that DynaST is readily applicable 
for universal exemplar-guided image generation
under both supervised and unsupervised settings.
%Moreover, we show that DynaST is adaptable to configurations of training objectives for both supervised and unsupervised scenarios, which makes it a unified solution for universal exemplar-guided image generation problems. 

We conduct extensive evaluations on three challenging tasks: 
pose-guided person image generation, edge-based face synthesis, 
and undistorted image style transfer. 
In all experiments, DynaST outperforms state-of-the-art 
exemplar-guided image generation models significantly, 
by up to 36.7\% in quantitative metrics, 
and achieves 
near-real-time inference efficiency,
with more than $2\times$ speedup compared
with previous state-of-the-art full-resolution matching solutions.

% Our contributions are  summarized as follows:
% \begin{itemize}
%     \item We introduce a novel multi-scale Transformer-based pipeline termed DynaST, 
%     which is the first dedicated Transformer-based approach
%     for exemplar-guided image generation. 
%     %This backbone enables DynaST to 
%     %model high-order feature dependency with its multi-layer architecture. 
%     This backbone enables DynaST to explore and rectify potential matching candidates more sufficiently with its multi-layer architecture and thereby models high-order feature dependency better. 
%     \item We propose a dynamic sparse attention mechanism 
%     in place of vanilla dense attention, so as
%     to capture optimal numbers of matching for different queries and establish efficacious full-resolution matching efficiently.
%     \item Thanks to the unified training objective, DynaST readily serves as a versatile model for various exemplar-guided image generation tasks in both supervised and unsupervised cases. 
%     Extensive qualitative and quantitative comparisons performed on multiple tasks and datasets 
%     verify the superiority of DynaST.  
% \end{itemize}

\section{Related Works}

\subsection{Exemplar-Guided Image Generation}
Exemplar-guided image generation has recently emerged as a 
popular task in the computer vision community. 
Park \textit{et al.}~\cite{park2019semantic} propose spatially-adaptive normalization (SPADE) module that generates normalization parameters based on given semantic information. 
In their work, exemplar images are fed to a VAE to encode the overall style and appearance, which guide the following generation procedure. 
Nevertheless, it is difficult to migrate local details in exemplar images due to such global transformation. 
Similar drawback also exists in works like~\cite{ma2017pose,wang2019example,albahar2019guided,zhu2019progressive,zheng2020example,tang2020xinggan,jing2020dynamic}. 
To enhance the generation of local textures, Ren \textit{et al.}~\cite{ren2022neural} introduce a neural texture extraction and distribution module. 
Recent attempts have also been made to introduce point-wise attention to exemplar-guided image generation and achieved superior results. 
For instance, Zhang \textit{et al.}~\cite{zhang2020cross} propose CoCosNet to learn the correspondence between input semantics and reference images. 
Zhan \textit{et al.}~\cite{zhan2021unbalanced} use unbalanced optimal transport to achieve the same goal. 
However, without any sparse mechanism, the quadratically increasing memory cost prevents these methods from learning fine-grained correspondence, which is important for synthesizing high-quality images. 
Although Zhou \textit{et al.}~\cite{zhou2021cocosnet} propose CoCosNet-v2 that is capable to learn full-resolution correspondence, the iterative global searching process by GRU makes a negative influence on the efficiency. 
DynaST in this paper brings the best of two worlds: it establishes matching at fine scales based on the dynamic sparse mechanism, which can generate images with high-quality local details while maintaining high efficiency simultaneously.

\subsection{Image-wise Matching}
Given a pair of images, image matching such as~\cite{jin2021image,li2020dual,liu2020extremely,sarlin2020superglue,truong2020glu,jiang2021cotr,sun2021loftr} aims to find pixel-wise correspondence leveraging local features, which is a fundamental problem in computer vision and is one related field to exemplar-guided image generation in this paper. 
The key difference is that cross-domain matching establishment, semantic map to the exemplar image, is required in the image synthesis problem unlike matching between two highly correlated images. 
This is also one major difference between the general cross-domain exemplar-guided image generation and reference-based image super-resolution~\cite{lu2021masa,zhang2019image,jiang2021robust,yang2020learning}. 

% Recently, Transformer is also brought into the area of image matching~\cite{jiang2021cotr,sun2021loftr} in favor of its global receptive field and strong ability to handle complex relationship. 
% Nevertheless, the Transformer in these works is still designed to densely take all tokens into account. 
% Furthermore, since all ground-truths are given as one-to-one pair, the dynamic nature of matching like those shown in Fig. \ref{fig:teaser} is not a major issue compared with that in the generation problem. 

\subsection{Efficient Transformer}
The full token-wise attention operation in standard Transformer~\cite{vaswani2017attention,dosovitskiy2020image} 
poses high requirements on memory and significantly increases computational cost. 
Thus, a lot of works are devoted to designing efficient attention mechanisms for Transformer or by extension,
graph-based methods~\cite{yang2020CVPR,yang2020NeurIPS}. 
On the one hand, some works rely on heuristic strategies to lead a current token only focus on those in a certain local context~\cite{child2019generating,dai2019transformer,li2019enhancing,beltagy2020longformer}. 
Recently, more strategies based on properties of images to boost the efficiency in vision Transformer are explored~\cite{Weihao22MetaFormer,Sucheng2022CVPR}.  
On the other hand, random sampling based Informer~\cite{zhou2021informer}, locality sensitive hashing based Reformer~\cite{kitaev2020reformer}, and approximated Softmax based Performer~\cite{choromanski2021rethinking} achieve lower complexity with a fine theoretical guarantee. 
Wang \textit{et al.}~\cite{wang2021kvt} only involve tokens with top $K$ attention scores for feature aggregation. 
Similar strategy is also adopted in~\cite{zhou2021cocosnet,zhan2021bi,tan2021efficient}. 
Although effective, it is not flexible enough to fix the number of attentive tokens, which fails to model the complex and changeable matching patterns in practice. 
%Rao \textit{et al.}~\cite{rao2021dynamicvit} propose dynamic token pooling to pruning useless tokens in vision tasks. 
Different from all these methods, the sparse mechanism in the attention module of this paper is based on prior knowledge in image matching, targeting at exemplar-guided image generation. 
%The dynamic mechanism is designed for attention pruning  and to capture complex matching patterns,  contributing to synthesizing high-quality results. 

\begin{figure*}[t]
  \centering
  \includegraphics[width=\textwidth]{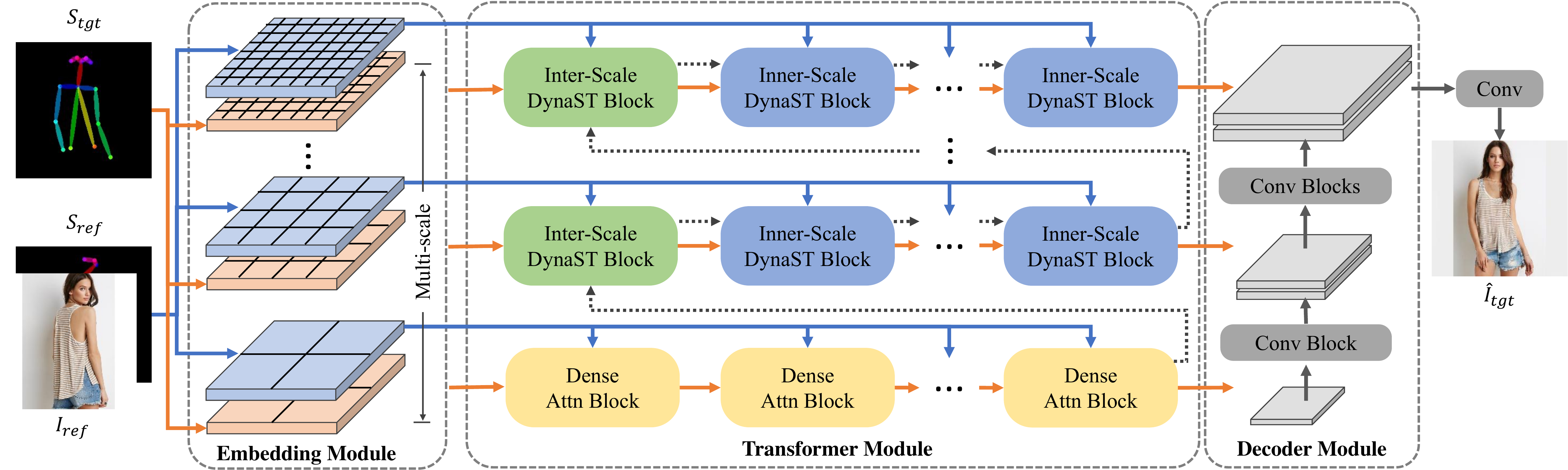}
%   \vspace{-0.8cm}
  \caption{DynaST overview. Solid arrows denote flows of features and dotted ones denote inheritance of attention maps. Yellow, green, and blue blocks in the middle adopt dense attention, inter-scale sparse attention, and inner-scale sparse attention respectively.}
%   \vspace{-0.6cm}
  \label{fig:overview}
\end{figure*}

\section{Dynamic Sparse Transformer}

In this section, we illustrate details
of the proposed DynaST model for exemplar-guided image generation.
The overview of DynaST is shown in Fig. \ref{fig:overview}. 
DynaST takes three images as input: a reference image $I_{ref}$, a corresponding semantic map $S_{ref}$ of $I_{ref}$ (\textit{e.g.}, a pose image or an edge map), and a target semantic map $S_{tgt}$.
It aims at synthesizing the image $\hat{I}_{tgt}$ with the target semantic information specified in $S_{tgt}$ and the appearance as well as the style in $I_{ref}$. 

The proposed DynaST consists of three parts.
The first is an \textit{embedding module} (Sec.~\ref{sec:3-1}), 
which is established by a set of multi-scale layers to extract and aggregate features at different levels. 
The second is a \textit{Transformer module} (Sec.~\ref{sec:3-2}), which restores features of target images with features of the semantic map as target and features of reference information as memory.
The last one is a lightweight \textit{decoder module} to synthesize final images, where the multi-scale features generated by the Transformer module are the input. 
The training objectives and supervised signals for the pipeline are described in Sec. \ref{sec:3-3}. 

\subsection{Embedding Module}
\label{sec:3-1}
Given an input semantic image $S_{tgt}$ and a reference image $I_{ref}$ along with its corresponding semantics $S_{ref}$, the embedding module produces a set of feature embedding, $F_{tgt}$ and $F_{ref}$.
%Different from the widely used vision Transformer architecture~\cite{dosovitskiy2020image} that only considers patch division at one single scale for downstream tasks, 
DynaST adopts a hierarchical patch embedding module as a multi-scale generative model, to enable the scale-wise cascaded matching process. The proposed embedding module is utilized to obtain rich features and contextual representation. Besides, the position embedding is also included to make the network aware of the positional information in the following matching process. 
%\textbf{Feature Embedding:}
Specifically, we use two independent sets of linear transformations: $\rm{E^i_{tgt}}$ and $\rm{E^i_{ref}}$, to obtain multi-scale patch embedding for target semantic map $S_{tgt}$ 
as well as
reference information $I_{ref}$ and $S_{ref}$, 
where  $i$ denotes the embedding of the $i$-th scale
with patch size $2^i\times2^i$. 
The features at the $i$-th scale, $F_{tgt}^i$ and $F_{ref}^i$, can then be written as:
\begin{equation}
\small
\begin{aligned}
    F^i_{tgt}={\rm X}([{\rm E^j_{tgt}}&(S_{tgt})'|0<j<M]),\\
    F^i_{ref}={\rm Y}([{\rm E^j_{ref}}([&S_{ref},I_{ref}])'|0<j<M]),
\end{aligned}
\end{equation}
where $0\leq i<M$, $M$ is the number of scales, notation~$'$ denotes the bilinear interpolation to unify the spacial dimension for scale $i$, $[\cdot]$ stands for the channel-wise concatenation, and ${\rm X}$ and ${\rm Y}$ are two MLPs consisting of two convolutional blocks for non-linear transformation. 
In other words, features of one specific scale aggregates multi-level patch embedding information, which provides rich features and contextual representations for the following matching and transformation steps. 
We also concatenate a learnable positional embedding to $F_{tgt}$ and $F_{ref}$ before computing their attention scores, denoted as $F^{i,pos}_{tgt}$ and $F^{i,pos}_{ref}$ at the $i$-th scale. 
The details can be found in the supplement. 

% \textbf{Position Embedding:} We concatenate positional embedding to $F_{tgt}$ and $F_{ref}$ before computing their attention scores. Thus the corresponding position embedding at the $i$-th scale, $F^{i,pos}_{tgt}$ and $F^{i,pos}_{ref}$ are:
% \begin{equation}
%     F^{i,pos}_{tgt}=[F^i_{tgt},pos_i],F^{i,pos}_{ref}=[F^i_{ref},pos_i],
% \end{equation}
% where $0\leq i<M$ and the position embedding for the coarsest scale $pos_{M-1}$ is a learnable tensor with the same spatial dimension as $F^{M-1}$. 
% Then, for upper levels ($0\leq i<M-1$), position encoding $pos_{i}$ is generated with learnable upsample-convolution-nonlinear blocks based on $pos_{i+1}$:
% \begin{equation}
%     pos_{i}=LReLU(Conv(Up(pos_{i+1}))),
% \end{equation}
% where $LReLU$ denotes leaky ReLU activation function.

\subsection{Transformer Module}
\label{sec:3-2}
The Transformer module is built for exemplar-guided image generation, by a set of dynamic sparse Transformer blocks (DynaST Block) and dense attention blocks (Dense Attn Block). The construction of these blocks consists of four steps, which are attention map computation, dynamic attention pruning, feature aggregation, and non-linear transformation.
In the attention map computation step, Dense Attn Blocks are used at the coarsest scale to construct matching at a low resolution, while DynaST Blocks are used at higher levels to infer high-resolution matching based on previous attention results. 
{At each higher scale, 
the first DynaST block adopts inter-scale sparse attention,
and is therefore named as Inter-Scale DynaST Block;
the remaining blocks adopt inner-scale sparse attention
and are thus named as Inner-Scale DynaST Blocks.
We will give details in the following sections.}

%which will be illustrated in the following parts. 
%Furthermore, DynaST Block can be further classified into \xw{Inter-Scale DynaST Block and Inner-Scale DynaST Block according to the different prior knowledge adopted for attention candidate selection, used as the first and the following Transformer blocks respectively at each scale, which will be illustrated in the following parts.}
%\xwc{too long, not clear, re-write}

\begin{figure}[!t]
    \includegraphics[width=\linewidth]{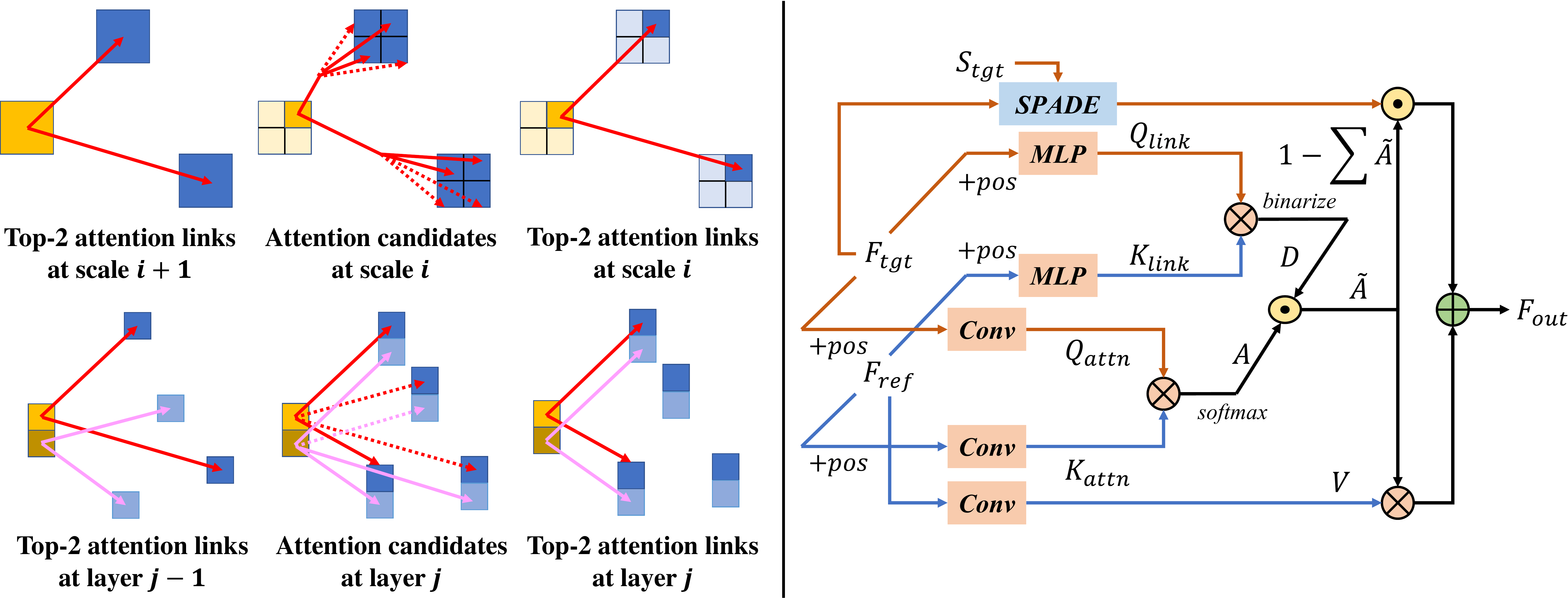}
    % \vspace{-0.6cm}
    \caption{Left: Intuition of inter-scale (up) and inner scale (bottom) sparse attention mechanism. Here $k=2$ is used for illustration. The yellow tokens represent queries in the target feature map and the blue ones represent tokens in the reference map. The dotted lines denote links masked by dynamic attention pruning. Right: Illustration of dynamic pruning and feature aggregation in DynaST blocks.}
    % \vspace{-0.6cm}
    \label{fig:intuition}
\end{figure}

\noindent\textbf{Attention Map Computation.}
Attention map computation is crucial to guide each target semantic feature point to focus on correct positions on reference feature maps and further restore features of target images. 
In DynaST, at scale $i$, the $j$-th DynaST block takes feature map produced by the $j-1$-th block $F^i_{tgt,j-1}$, the reference feature map $F^i_{ref}$, and attention map of previous DynaST block as input to compute attention scores. 
Note that $F^i_{tgt,0}$ is defined as $F^i_{tgt}$ from the multi-scale patch embedding layer.
At the coarsest level $i=M-1$, Dense Attn Blocks with vanilla attention are used to derive attention scores:
\begin{equation}
\small
    A_j^{M-1}={\rm softmax}(\tau\alpha(\overline{F^{M-1,pos}_{tgt,j-1}})\beta(\overline{F^{M-1,pos}_{ref}})^\top),\label{eq:dense_attn}
\end{equation}
where $\alpha$ and $\beta$ are implemented as two $1\times1$ convolutional kernels, $\tau$ is a hyper-parameter controlling the smoothness of attention distribution, and $\overline{x}$ denotes channel-independent instance normalization~\cite{ulyanov2016instance}. 

Then, at finer level $i<M-1$, for the first DynaST block ($j=1$), \emph{inter-scale sparse attention} is proposed to compute the attention map at this layer:
\begin{equation}
\small
    A_1^i\!=\!{\rm softmax}(\tau\alpha(\overline{F^{i,pos}_{tgt,0}})\beta({\rm TopK}_{A_{-1}^{i+1}}(\overline{F^{i,pos}_{ref}}))^\top),\label{eq:inter_attn}
\end{equation}
where ${\rm TopK}_{A_{-1}^{i+1}}$ denotes that matching candidates for a point in $F^{i,pos}_{tgt,0}$ come from those with top $k$ large scores in the attention map of the last block in the previous scale.
Note that a point in the previous scale would be divided into four in the current one. 
Therefore, there are $k\times4$ matching candidates for attention map computation at this layer. 

For the following blocks ($j>1$), {inner-scale sparse attention} is performed to refine the attention matching at the current scale, based on the prior knowledge that the matching offset in a local area is likely to be the same~\cite{barnes2009patchmatch}:
\begin{equation}
\small
    A_j^i={\rm softmax}(\tau\alpha(\overline{F^{i,pos}_{tgt,j-1}})\beta({\rm\mathcal{N}}({\rm TopK}_{A_{j-1}^i}(\overline{F^{i,pos}_{ref}})))^\top),\label{eq:inner_attn}
\end{equation}
%where notation $\mathcal{N}$ is the operation to derive the neighboring points of the previous matching results of the neighbors for one target point\xwc{need to be simplified}. 
where notation $\mathcal{N}$ is the operation to derive the points with the same matching offsets as the neighbors for one target point. 
%For example, for one target point at the current layer, it would take the left neighboring points of matching results of its right neighbor as candidates \xwc{one 'of' in one phrase}. 
For example, for one target point at the current layer, it firstly finds matching results of its right neighbor and then takes the left neighbors of these points as candidates. 
We define the up, bottom, left, and right points of one position plus the current point itself as the neighboring points in this paper. 
In this way, the number of matching candidates for each target point in the inner-scale sparse attention layer is $
k\times5$. 
The intuitions of inter/inner-scale sparse attention layer are illustrated in Fig. \ref{fig:intuition} Left. %Thus, the block utilizes 

\noindent\textbf{Dynamic Attention Pruning.}
Considering that not all the matching candidates in attention modules are necessary for feature aggregation, we propose dynamic attention pruning to decide whether an attention link between a point in the target map and that in the reference map is useful. 
To this end, we use two MLPs $\Omega$ and $\Phi$ to transform $F_{tgt}$ and $F_{ref}$ into a common feature space. 
Then, a sign function is applied on the inner product of the transformed results to obtain the decision $D$ for each attention link:
\begin{equation}
\small
\begin{aligned}
    P_j^i&={\rm \Omega}(F_{tgt,j-1}^i){\rm \Phi}(F_{ref}^i)^\top\\
    D_j^i&=\left\{
    \begin{aligned}
    &1,\quad P_j^i>0, \\
    &0,\quad otherwise
    \end{aligned}
    \right. 
\end{aligned}
\end{equation}
Note that in the above function, the sign function introduces obstacles for gradient-based optimization in training. 
To tackle this issue, we alternatively take gradients from $sigmoid$ function during the backward propagation:
\begin{equation}
\small
    \frac{{\rm d}D_j^i}{{\rm d}P_j^i}=\frac{\exp(-P_j^i)}{(1+\exp(-P_j^i))^2}.
\end{equation}

Thus, attention maps after the dynamic pruning operation are derived by:
\begin{equation}
\small
    \tilde{A_j^i}=D_j^i\odot A_j^i,
\end{equation}
where $\odot$ represents the element-wise multiplication. 

\noindent\textbf{Feature Aggregation.}
One straightforward way to conduct feature aggregation is to use pruned attention matrix $\tilde{A_j^i}$ directly to perform weighted summation over reference features. 
However, due to the pruning operation, the sum of attention weights for one target point to all reference feature points is no longer guaranteed to be $1$, which would result in unbalanced magnitudes in feature aggregation. 
For example, in the most extreme case, a target point would be untraceable in the reference image and attention decisions would be all $0$ for this point. 
Then the aggregated features for this point would also be all $0$, which impedes the synthesis of a plausible image. 
To alleviate this problem, we use features restored by a SPADE block $SP$~\cite{park2019semantic} to compensate for the masked part by dynamic attention pruning:
%\footnote{Multi-head attention is adopted to concatenate features produced by different attention heads. The notation is omitted for simplicity.}: 
\begin{equation}
\small
\begin{aligned}
    F_{out}=(1-\sum\tilde{A_j^i})\odot &{\rm SP}(F^i_{tgt,j-1},S_{tgt})+\tilde{A_j^i}{\rm \eta}(F_{ref}^i),\\
    {F^i_{tgt,j}}'={\rm Norm}&(F_{out}+F^i_{tgt,j-1}),
\end{aligned}
\end{equation}
where the summation is along the dimension of reference feature points, $\eta$ is another $1\times1$ convolutional kernel, and $Norm$ denotes the layer normalization. 
Key steps for feature aggregation in DynaST blocks are shown in Fig. \ref{fig:intuition} Right. 

\noindent\textbf{Non-Linear Transformation.}
Finally, following the standard Transformer architecture, a residual block is added at the end of the DynaST block for non-linear transformation:
\begin{equation}
\small
    F^i_{tgt,j}={\rm Norm}({F^i_{tgt,j}}'+{\rm Conv}({\rm ReLU}({\rm Conv}({F^i_{tgt,j}}')))).
\end{equation}

\subsection{Training Objectives}
\label{sec:3-3}

DynaST is a universal framework for exemplar-guided image generation, which is compatible for objectives of both supervised and unsupervised tasks. 
The overall training objective consists of two parts by default: task-specific loss $\mathcal{L}_t$ and matching loss $\mathcal{L}_m$.

\noindent\textbf{Task-Specific Loss.}
Firstly, task-specific loss $\mathcal{L}_t$ targets at the task itself and is flexible for different forms of loss functions in our model. 
Typically, for supervised tasks like pose-guided person image generation, the objective is defined as the MSE between generated images $\hat{I}_{tgt}$ and ground truths $I_{tgt}$ in original image space and perceptual feature space plus an adversarial loss term:
\begin{equation}
\small
\begin{aligned}
    &\mathcal{L}_t=\Vert \hat{I}_{tgt}-I_{tgt}\Vert_2^2+\sum_i\lambda_i\Vert {\rm \phi_i}(\hat{I}_{tgt})-{\rm \phi_i}(I_{tgt})\Vert_2^2+\\&\lambda_{adv}\max\{\log {\rm Dis}(S_{tgt},I_{tgt})+\log (1-{\rm Dis}(S_{tgt},\hat{I}_{tgt}))\},
\end{aligned}
\end{equation}
where $\phi$ denotes a pretrained feature extractor (\textit{e.g.}, VGG-19), subscript $i$ specifies which layer features come from, $\lambda_i$ controls the weight of each layer, $Dis$ represents a discriminator to be trained alternatively with the generator, and $\lambda_{adv}$ is the weight of the adversarial term~\cite{albahar2019guided}. 
For another example, style transfer is an unsupervised task, whose loss function can be written as:
\begin{equation}
\small
\begin{aligned}
    \mathcal{L}_t=l_c+\lambda_s l_s,\quad
    l_c=\Vert &{\rm \phi_{4\_1}}(I_{cs})-{\rm \phi_{4\_1}}(I_{c})\Vert_2^2\\
    l_s=\sum_{i=1}^4(\Vert\mu({\rm \phi_{i\_1}}(I_{cs}))-\mu({\rm \phi_{i\_1}}(I_s)))\Vert_2^2&+\Vert\sigma({\rm \phi_{i\_1}}(I_{cs}))-\sigma({\rm \phi_{i\_1}}(I_s)))\Vert_2^2,
\end{aligned}
\end{equation}
where $\phi_{x\_1}$ aims to extract features of \texttt{ReLU\_x-1} layer of a VGG-19 network pretrained on ImageNet~\cite{deng2009imagenet}, and $\mu$ and $\sigma$ denote mean and standard deviation of each feature channel respectively~\cite{johnson2016perceptual,huang2017arbitrary}. 
In DynaST, content images $I_c$ and style images $I_s$ are input to the embedding modules $E_{tgt}$ and $E_{ref}$ respectively, and style transfer images $I_{cs}$ are the framework output $\hat{I}_{tgt}$.
%\xwc{Don't understand here. What is ``input as''?}

\noindent\textbf{Matching Loss.}
To provide a more direct supervision signal for matching modules and dynamic pruning modules to produce proper attention maps, we introduce a matching loss $\mathcal{L}_m$ that uses the output attention maps to warp the reference images and measures the task-specific loss produced by the warped images. 
To be specific, we denote the correlation maps, results of Eq. \ref{eq:dense_attn}, \ref{eq:inter_attn}, and \ref{eq:inner_attn} before $softmax$, by $C_j^i$. 
The warp matrices $W_j^i$ are derived taking attention decision $D_j^i$ into consideration:
\begin{equation}
\small
    W_j^i=\frac{D_j^i\odot\exp(C_j^i)}{\sum\{D_j^i\odot\exp(C_j^i)\}+\epsilon},\label{eq:warp}
\end{equation}
where the summation is over reference feature points and $\epsilon$ is a small constant for numerical stability. 
Then, the warped reference images are derived by:
\begin{equation}
\small
    \hat{I}^i_{warp,j}=W_j^iI_{ref}',
\end{equation}
where $I_{ref}'$ denotes the resized version of the reference images to keep dimension scales of $W_j^i$ and $I_{ref}$ the same. 
Matching loss is defined by the MSE:
\begin{equation}
\small
    \mathcal{L}_m=\sum_{i}\sum_{j}\Vert \hat{I}^i_{warp,j}-I'_{tgt}\Vert_2^2.\label{eq:match} 
\end{equation}

Finally, the overall objective is given by a weighted sum of $\mathcal{L}_t$ and $\mathcal{L}_m$:
\begin{equation}
\small
    \mathcal{L}=\mathcal{L}_t+\lambda_m\mathcal{L}_m,
\end{equation}
with $\lambda_m$ controlling the weight of term $\mathcal{L}_m$.

\section{Experiments}

\textbf{Implementation Details.}
For all the experiments, DynaST are trained under $256\times256$ resolution. $4$ different scales are set to be $32$, $64$, $128$ and $256$, where dimensions of the corresponding feature channel are $512$, $256$, $128$ and $64$ respectively. 
%The networks are initialized under normal distribution with $\mu=0$ and $\sigma=0.02$. 
Each level of Transformation module is built by 2 blocks, where on the coarsest scale there are $2$ \textit{Dense Attn Blocks} and on each upper level there is $1$ \textit{Inter-Scale DynaST Block} and $1$ \textit{Inner-Scale DynaST Block}. 
% Other training settings are the same as~\cite{zhang2020cross}. 
For supervised tasks, hyper-parameters $\lambda_m$ and $\lambda_{adv}$ are set as $100$ and $10$, respectively. 
% Perceptual loss is measured on a pretrained VGG-19 encoder. 
% Features on \texttt{ReLU-x\_2} layers ($1\leq x\leq5$) are used with weights $0.3125$, $0.625$, $1.25$, $2.5$ and $10$ for each layer. 
For style transfer, $\lambda_m$ and $\lambda_s$ are set as $1$ and $3$, respectively. 
Matching loss defined in Eq. \ref{eq:match} is not adopted here. 
%In style transfer, the \textit{SPADE} branch in DynaST blocks is removed and Eq. \ref{eq:warp} is alternatively adopted to calculate attention scores. 
The smoothness parameter $\tau$ is set as $100$ and $k=4$ is used by default when selecting attention candidates. 
Training is done on 8 Tesla V100 GPUs with batch size $32$. 

\begin{figure*}[t]
  \centering
  \includegraphics[width=\textwidth]{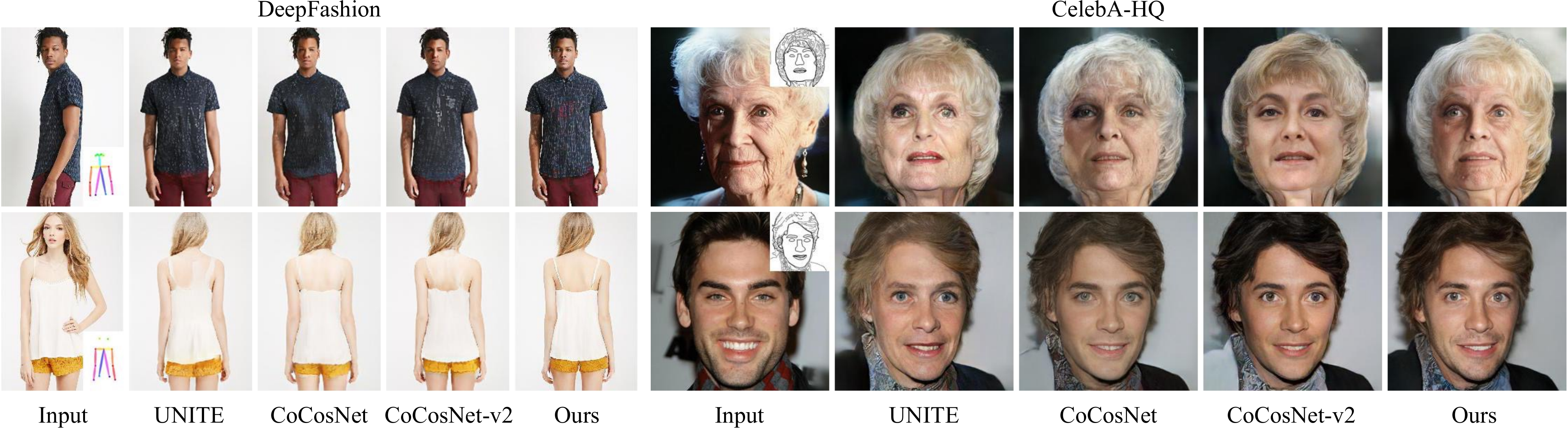}
%   \vspace{-0.7cm}
  \caption{Comparisons with state-of-the-art exemplar-guided image generation methods on \textit{DeepFashion} and \textit{CelebA-HQ} datasets.}
%   \vspace{-0.6cm}
  \label{fig:compare_qualitative}
\end{figure*}

\textbf{Datasets.}
For the pose-guided person image generation and the edge-guided face generation tasks, \textit{DeepFashion}~\cite{liu2016deepfashion} and \textit{CelebA-HQ}~\cite{liu2015deep} datasets are used. 
Splits of training and validation sets and policies on retrieval of input-exemplar image pairs are consistent with those in~\cite{zhang2020cross}. 
For style transfer, following the common settings, \textit{MS-COCO}~\cite{lin2014microsoft} and \textit{WikiArt}~\cite{phillips2011wiki} are adopted as content and style image sets for training respectively. 
During training, all the images are resized to $512\times512$ and then randomly cropped to $256\times256$. 
Inference results under $512\times512$ resolution are reported in this paper. 

\subsection{Supervised Tasks}

\textbf{Comparison with Other Methods.}
On the pose-guided person image generation and the edge-guided face generation problems, we mainly compare our method with three state-of-the-art attention-based exemplar-guided image generation methods, including UNITE~\cite{zhan2021unbalanced}, CoCosNet~\cite{zhang2020cross}, and CoCosNet-v2~\cite{zhou2021cocosnet}. 
The attention matching in UNITE and CoCosNet is limited on a relatively coarse scale ($64\times64$) by the quadratic memory cost of the dense attention operation. 
Thus, as shown in Fig.~\ref{fig:compare_qualitative}, some detailed textures are not good enough, \textit{e.g.}, cloth and face details in the 1st row of comparisons on \textit{DeepFashion} and the beard in the 2nd row of comparisons on \textit{CelebA-HQ}. 
CoCosNet-v2 leverages the Conv-GRU module to predict correspondence at fine scales, where noisy correspondences would be added easily due to the large search space under high resolutions. And there is no pruning to mask irrelevant matching in CoCosNet-v2, which may lead to some artifacts, \textit{e.g.}, straps in the 2nd row of comparisons on \textit{DeepFashion}. 
As shown in the last column of each example, our method addresses these problems successfully with the proposed dynamic sparse attention based Transformer model, which generates more high-quality results. 

\begin{figure}[t]
  \centering
  \includegraphics[width=0.6\linewidth]{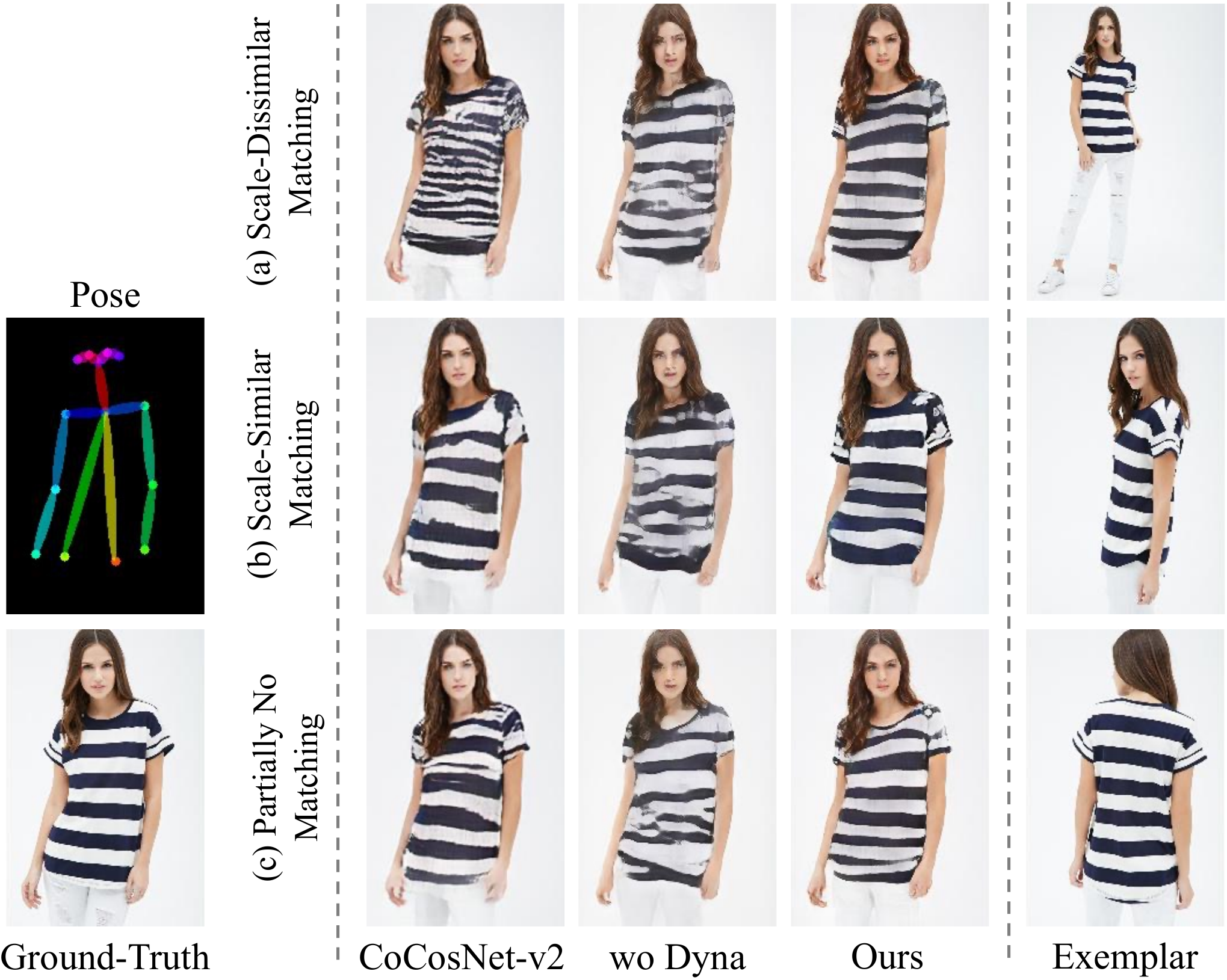}
%   \vspace{-0.3cm}
  \caption{Challenging scenarios involving
  (a) scale-dissimilar matching, 
  (b) scale-similar matching, 
  and (c) partially no matching,
  where the static matching scheme inevitably
  fails.
  The proposed DynaST, thanks to its dedicated
  scheme for handling dynamic numbers of matching,
  may well handle all such cases and yield
  visually plausible results.}
%   \vspace{-0.4cm}
  \label{fig:compare_dyna}
\end{figure}

Notably, one major difference between our DynaST and CoCosNet-v2 is that DynaST uses a dynamic number of matching points for feature aggregation, while CoCosNet-v2 only considers a fixed number of candidates, which fails to account for the dynamic property of matching in different cases. 
One illustrative example is shown in Fig. \ref{fig:compare_dyna}, where results by CoCosNet-v2 are less robust to cases when scales of input and exemplar are different, since one location must select a fixed number of matching points. 
When there are less informative correspondences, noises are inevitably introduced. 
By comparison, dynamic pruning involved in DynaST is more competent to handle such scale variation robustly. 

\begin{table}[!t]
\scriptsize
\centering
    \begin{tabular}{c|cccc|cccc}
        \toprule
        \multicolumn{1}{c}{\multirow{2}{*}{Method}} & \multicolumn{4}{c}{DeepFashion} & \multicolumn{2}{c}{DeepFashion} & \multicolumn{2}{c}{CelebA-HQ} \\
        \cmidrule(r){2-5}\cmidrule(r){6-7}\cmidrule(r){8-9}
        & L1$\downarrow$ & PSNR$\uparrow$ & SSIM$\uparrow$ & Time$\downarrow$ & FID$\downarrow$ & SWD$\downarrow$ & FID$\downarrow$ & SWD$\downarrow$ \\
        \midrule
        Pix2PixHD~\cite{wang2018high} & - & - & - & - & 25.2 & 16.4 & 62.7 & 43.3 \\
        SPADE~\cite{park2019semantic} & - & - & - & - & 36.2 & 27.8 & 31.5 & 26.9 \\
        MUNIT~\cite{huang2018multimodal} & - & - & - & - & 74.0 & 46.2 & 56.8 & 40.8 \\
        EGSC-IT~\cite{ma2018exemplar} & - & - & - & - & 29.0 & 39.1 & 29.5 & 23.8 \\
        UNITE~\cite{zhan2021unbalanced} & 13.1 & \underline{16.7} & \underline{13.2} & 14.9 & 13.1 & \underline{16.7} & \underline{13.2} & 14.9 \\
        CoCosNet~\cite{zhang2020cross} & 0.067 & 18.48 & 0.80 & \underline{11.5} & 14.4 & 17.2 & 14.3 & 15.2 \\
        CoCosNet-v2~\cite{zhou2021cocosnet} & \underline{0.064} & 18.24 & 0.80 & 21.7 & \underline{13.0} & \underline{16.7} & \underline{13.2} & \underline{14.0} \\
        \midrule
        Ours-32 & 0.077 & 18.12 & 0.73 & 5.53 & 8.55 & 15.4 & 16.0 & 17.8 \\
        Ours-64 & 0.063 & 18.22 & 0.78 & 7.24 & 8.50 & 12.8 & 13.1 & 13.1 \\
        Ours-128 & 0.061 & 19.13 & 0.82 & 8.43 & 8.41 & 12.9 & 12.3 & 12.7 \\
        Ours wo Inner & 0.064 & 18.30 & 0.82 & 9.45 & 8.88 & 12.0 & 14.7 & 17.2 \\
        Ours wo Dyna & 0.063 & 19.04 & 0.81 & 9.26 & 9.32 & 21.8 & 15.3 & 19.0 \\
        \midrule
        Ours & \textbf{0.054} & \textbf{19.25} & \textbf{0.83} & 9.63 & \textbf{8.36} & \textbf{11.8} & \textbf{12.0} & \textbf{12.4} \\
        \bottomrule
    \end{tabular}
    \caption{Left: Quantitative metrics on the quality and efficiency of matching establishment. Results are measured by comparing warped results with the ground truth on DeepFashion dataset. Running time for one sample ($\times10^{-2}$ sec.) is shown here. Right: Quantitative metrics of image quality on two datasets.}
    % \vspace{-0.8cm}
    \label{tab:sota}
\end{table}

\begin{figure*}[t]
  \centering
  \includegraphics[width=\textwidth]{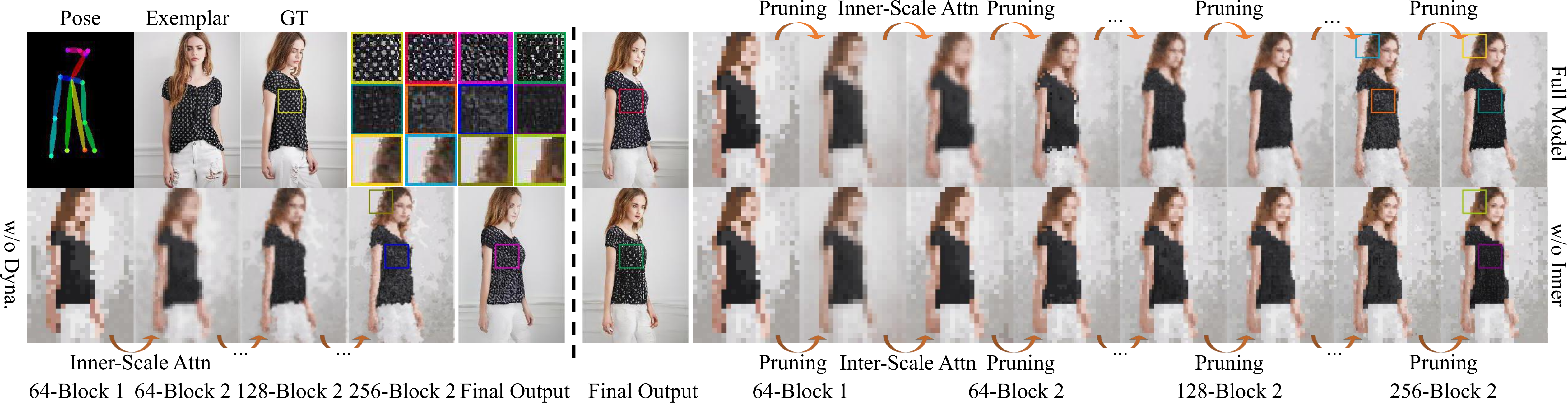}
%   \vspace{-0.8cm}
  \caption{Ablation studies on dynamic pruning and inner-scale sparse attention. Intermediate warping results using attention maps are visualized to demonstrate the evolving of input-exemplar matching. Zoom-in for better visualization.}
%   \vspace{-0.9cm}
  \label{fig:compare_warp}
\end{figure*}

To further illustrate the advantage of our method on matching construction, we use the attention matrix derived by each method to warp the exemplar image and report the warped results. 
As shown in Fig. \ref{fig:compare_warp}, results by UNITE and CoCosNet are blurry due to low-resolution matching, and results by CoCosNet-v2 contain too much noise. 
Compared with the above methods, our method can generate matching under the full resolution with the best quality. 

Quantitatively, leveraging the paired samples in \textit{DeepFashion} dataset, we compare warped results with ground-truth images and show the average L1 loss, PSNR, and SSIM scores in Tab. \ref{tab:sota} Left, where our method performs the best. 
The time needed for generating matching and synthesizing results for one sample is also included, measured on a single Nvidia 3090 GPU by averaging 1000 samples. 
Dense attention mechanism and iterative solving of optimal transport problem in CoCosNet and UNITE respectively leave a high computational burden. 
Recurrent prediction under full resolution in CoCosNet-v2 increases the latency further. 
Different from previous approaches, the efficient dynamic sparse attention operation in DynaST makes it achieve the most satisfactory computational speed while generating the best matching results impressively. 

\begin{table}[!t]
\scriptsize
\centering
    \begin{tabular}{cm{1.35cm}<{\centering}m{1.35cm}<{\centering}m{1.35cm}<{\centering}m{1.35cm}<{\centering}m{1.35cm}<{\centering}m{1.35cm}<{\centering}}
        \toprule
        \multicolumn{1}{c}{\multirow{2}{*}{Method}} & \multicolumn{3}{c}{DeepFashion} & \multicolumn{3}{c}{CelebA-HQ} \\
        \cmidrule(r){2-4}\cmidrule(r){5-7}
        & Sem.$\uparrow$ & Col.$\uparrow$ & Tex.$\uparrow$ & Sem.$\uparrow$ & Col.$\uparrow$ & Tex.$\uparrow$ \\
        \midrule
        Pix2PixHD & 0.943 & NA & NA & 0.914 & NA & NA \\
        SPADE & 0.936 & 0.943 & 0.904 & 0.922 & 0.955 & 0.927 \\
        MUNIT & 0.910 & 0.893 & 0.861 & 0.848 & 0.939 & 0.884 \\
        EGSC-IT & 0.942 & 0.945 & 0.916 & 0.915 & 0.965 & 0.942 \\
        UNITE & 0.957 & 0.973 & 0.930 & \textbf{0.952} & 0.966 & 0.950 \\
        CoCosNet & \underline{0.968} & \textbf{0.982} & \textbf{0.958} & \underline{0.949} & \underline{0.977} & \underline{0.958} \\
        CoCosNet-v2 & 0.959 & \underline{0.974} & 0.925 & 0.948 & 0.975 & 0.954 \\
        \midrule
        Ours & \textbf{0.975} & \underline{0.974} & \underline{0.937} & \textbf{0.952} & \textbf{0.980} & \textbf{0.959} \\
        \bottomrule
    \end{tabular}
    \caption{Quantitative metrics of semantic (Sem.), color (Col.), and texture (Tex.) consistency on two datasets compared with state-of-the-art image synthesis methods.}
    % \vspace{-0.75cm}
    \label{tab:consist}
\end{table}

\begin{figure*}[t]
  \centering
  \includegraphics[width=\textwidth]{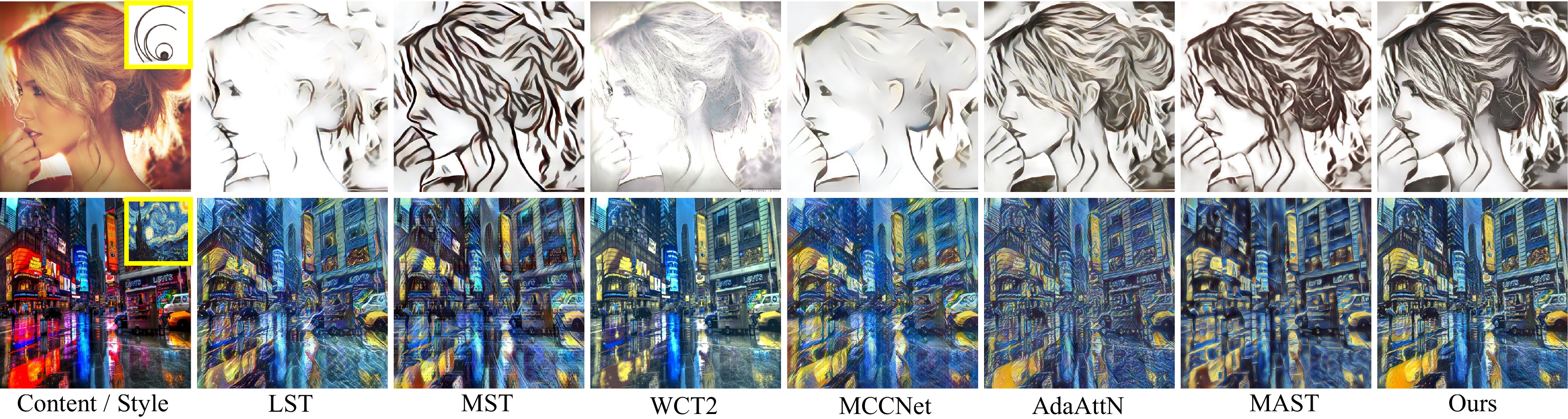}
%   \vspace{-0.9cm}
  \caption{Comparisons with state-of-the-art undistorted style transfer methods.}
%   \vspace{-0.9cm}
  \label{fig:compare_style}
\end{figure*}

We report the widely-used FID~\cite{heusel2017gans} and SWD~\cite{lee2019sliced} metrics to reflect the distance of feature distributions between generated samples and real images, following Zhang \textit{et al.}~\cite{zhang2020cross} in Tab. \ref{tab:sota} Right.
Our method outperforms previous ones significantly under both datasets, suggesting the best quality of results. 
On the other hand, we show the semantic, color, and texture consistency in Tab. \ref{tab:consist}, also under the same setting as~\cite{zhang2020cross}. 
% Semantic consistency is measured against input semantics, by taking the average cosine distance over the $ReLU-3\_2$, $ReLU-4\_2$, and $ReLU-5\_2$ layers of a pretrained VGG network~\cite{simonyan2014very}, while color and texture consistency is against exemplar images, using the shallower $ReLU-1\_2$ and $ReLU-2\_2$ layers respectively. 
%Semantic consistency is measured against input semantics, by taking the average cosine distance over deep $3$ layers of a pretrained VGG network~\cite{simonyan2014very}, while color and texture consistency is against exemplar images, using the shallow $2$ layers respectively. 
Results in Tab. \ref{tab:consist} demonstrate that our method achieves competent semantic restoration and style migration performance.

\textbf{Ablation Study.}
We conduct ablation studies on the two core ideas in this paper: dynamic pruning and sparse attention. 
Based on the full model, we (1) remove the dynamic pruning mechanism, with the corresponding result denoted as \textit{Ours wo Dyna}; (2) replace all inner-scale DynaST blocks with inter-scale ones, denoted as \textit{Ours wo Inner}; and (3) remove inter-scale sparse attention layers at different scales and only use attention up to a coarse scale instead of full resolution, denoted as \textit{Ours-$x$}, where $x\in\{32,64,128\}$ representing the highest resolution used for matching. 
Evaluations on warped results and final results mentioned above are repeated using the resulting models. 
The quantitative results in Tab. \ref{tab:sota} indicate that the incomplete models lead to inferior results. 

Qualitatively, we visualize the intermediate warping results using attention maps in Fig. \ref{fig:compare_warp}, to demonstrate how the matching results are evolving through multi-layer dynamic pruning and inter/inner-scale attention. 
For one thing, dynamic pruning is capable of suppressing noisy matching and helps generate a clearer view. 
The example in Fig. \ref{fig:compare_dyna} also demonstrates the importance of dynamic pruning for the robustness to handle scale variation. 
For another, replacing inner-scale DynaST blocks may lead to some artifacts such as checkerboard, as shown in Fig. \ref{fig:compare_warp}, due to the missing of local refinement. 
Removing them would have negative impacts on local details like areas of hair and cloth. 

\begin{wrapfigure}{r}{0.5\linewidth}
  \centering
%   \vspace{-0.9cm}
  \includegraphics[width=0.5\textwidth]{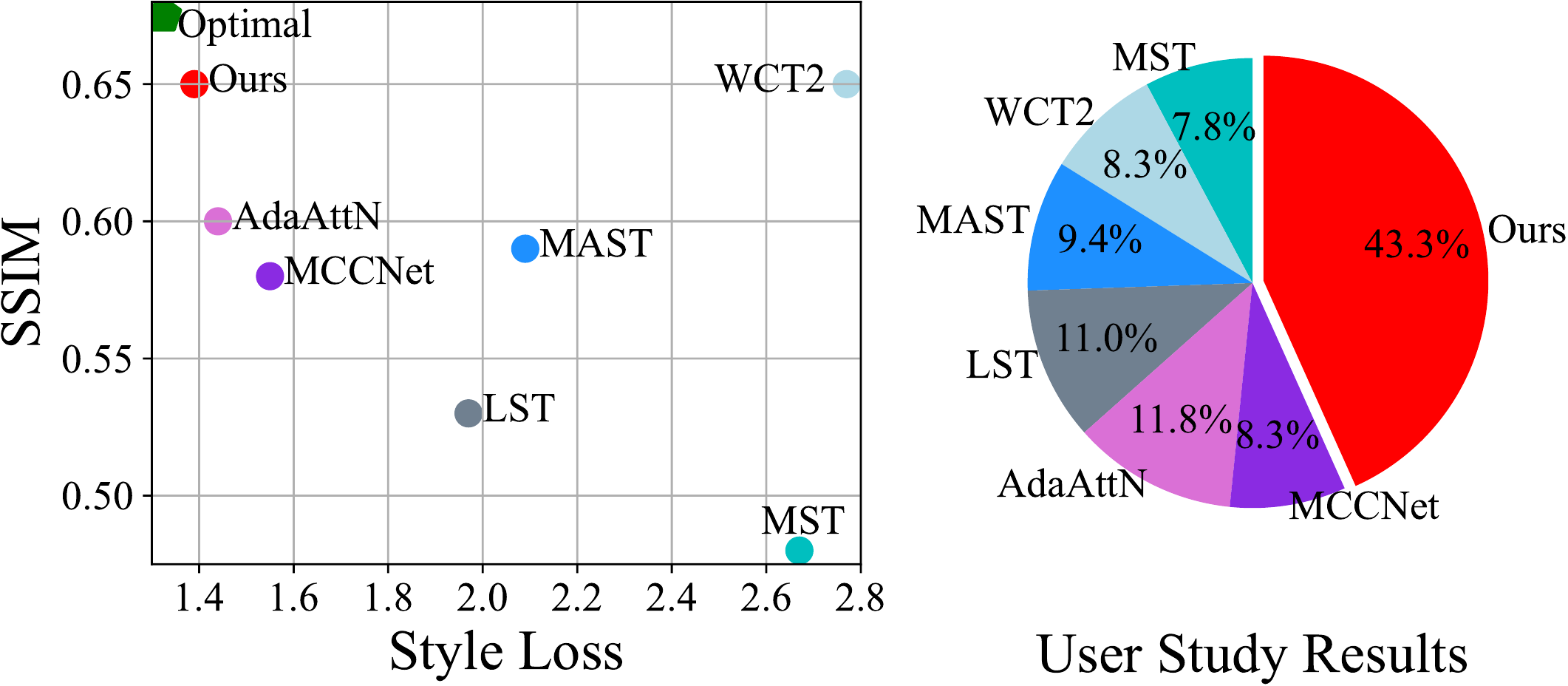}
%   \vspace{-0.9cm}
  \caption{Left: visualization of content SSIM and style loss of different undistorted style transfer methods; Right: distribution of user preference.}
%   \vspace{-0.8cm}
  \label{fig:quantitative_style}
\end{wrapfigure}

\subsection{Undistorted Image Style Transfer}
The full resolution matching mechanism makes DynaST well suitable to generate undistorted style transfer results. 
In order to demonstrate such advantage, we compare our DynaST with 6 state-of-the-art style transfer methods with the same or similar goal, including LST~\cite{li2019learning}, MST~\cite{zhang2019multimodal}, WCT2~\cite{yoo2019photorealistic}, MCCNet~\cite{deng2020arbitrary}, AdaAttN~\cite{liu2021adaattn}, and MAST~\cite{huo2021manifold}. 
As shown in Fig. \ref{fig:compare_style}, compared with the photorealistic style transfer method WCT2, our results migrate global and local style patterns better, while compared with other methods, our results preserve texture details of content images best, \textit{e.g.}, hair in the 1st row. 
In particular, our method is capable of dealing with complicated scenes like the 2nd row without distortion, where other methods fail. 
We also visualize \emph{SSIM scores} with content images and \emph{style loss} against style images in Fig. \ref{fig:quantitative_style} Left, using the test dataset in~\cite{yoo2019photorealistic}. 
It is demonstrated that DynaST, as the first full-resolution matching based method in style transfer, can even achieve comparable content preservation ability with photorealistic style transfer methods, while significantly improving the stylization effects. 
Furthermore, we conduct a \emph{user study} with the same test dataset to reflect user preference. 
There are $155$ users involved and $12$ content-style pairs are shown to each randomly. 
They are invited to select their favorite one for each pair among results by the 7 methods. 
We receive $1860$ votes in total and the preference distribution is shown in Fig. \ref{fig:quantitative_style} Right, where our preference score outperforms others significantly. 
In this way, both qualitative and quantitative comparisons demonstrate the superiority of DynaST. 

\section{Conclusion}
In this paper, 
we introduce a novel
multi-scale Transformer model,
DynaST,
to account for
dynamic sparse attention and
construct fine-level matching in exemplar-guided image generation tasks.
DynaST
comes with a unified training objective,
making it a versatile model 
for various exemplar-guided image generation tasks
under both supervised and unsupervised settings.
Extensive evaluations on multiple benchmarks demonstrate that the proposed DynaST outperforms previous state-of-the-art methods, on both the matching quality and the running efficiency. 

\section*{Acknowledgement}
This research is supported by the National Research Foundation Singapore under its AI Singapore Programme (Award Number: AISG2-RP-2021-023), and NUS Faculty Research Committee Grant (WBS: A-0009440-00-00).
\clearpage
\bibliographystyle{splncs04}
\bibliography{DynaST}

\clearpage
\appendix

\section{Position Embedding}
We concatenate positional embedding 
to $F_{tgt}$ and $F_{ref}$ before computing their attention scores:
\begin{equation}
    F^{i,pos}_{tgt}=[F^i_{tgt},{\rm pos_i}],F^{i,pos}_{ref}=[F^i_{ref},{\rm pos_i}],
\end{equation}
where $0\leq i<M$. The position embedding for the 
coarsest scale $pos_{M-1}$ is a learnable tensor with the same spatial dimension as $F^{M-1}$. 
For upper levels ($0\leq i<M-1$), position encoding ${\rm pos_{i}}$ is therefore generated with learnable upsample-convolution-nonlinear blocks based on ${\rm pos_{i+1}}$:
\begin{equation}
    {\rm pos_{i}}={\rm LReLU}({\rm Conv}({\rm Up}({\rm pos_{i+1}}))),
\end{equation}
where ${\rm LReLU}$ denotes leaky ReLU activation function.

\section{More Results}

\begin{figure*}[t]
  \centering
  \includegraphics[width=\textwidth]{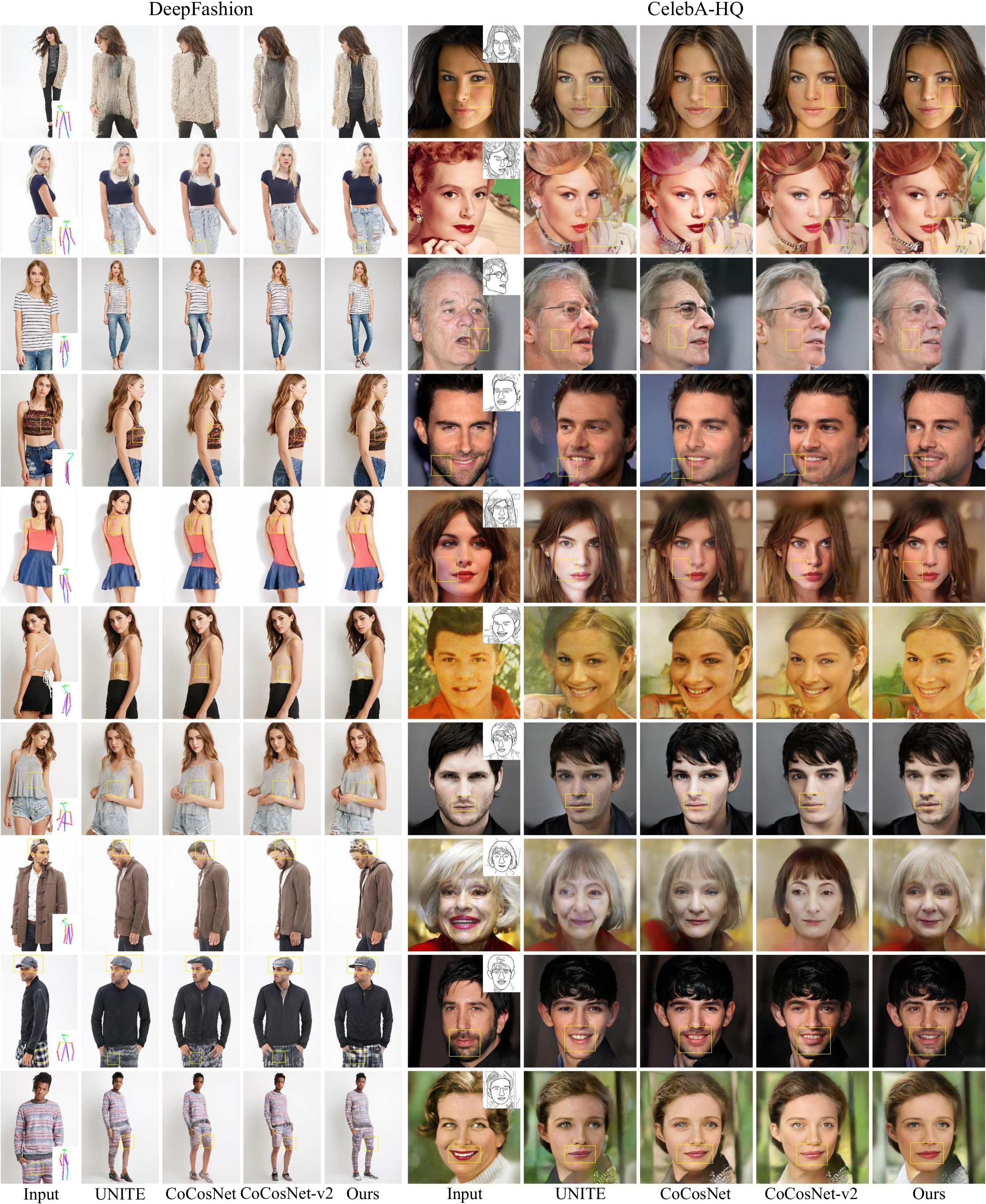}
%   \vspace{-0.7cm}
  \caption{More comparisons with state-of-the-art exemplar-guided image generation methods on two datasets.}
  \label{fig:compare_qualitative_supp}
\end{figure*}

\noindent\textbf{Supervised Tasks.} We provide more examples
to better demonstrate advantages of our DynaST over state-of-the-art exemplar-guided 
image generation methods, 
including UNITE~\cite{zhan2021unbalanced}, CoCosNet~\cite{zhang2020cross}, and CoCosNet-v2~\cite{zhou2021cocosnet}. 
On one hand, as shown in Fig. \ref{fig:compare_qualitative_supp} Left, 
DynaST generates better local details 
compared with other methods in pose-guided person image generation task 
on the \textit{DeepFashion} dataset, \textit{e.g.}, cloth appearances 
in the 1st, 5th, 6th, and 7th rows, hats in the 2nd, 8th, and 9th 
rows, and cloth textures in the 4th and 10th rows. 
In particular, when there is a scale variance 
between exemplar and target images like the 3rd row, 
DynaST yields the best cloth-appearances and -textures
restoration results, due to its 
dynamic attention mechanism. 
On the other hand, in edge-based face synthesis 
on the \textit{CelebA-HQ} dataset shown in 
Fig.~\ref{fig:compare_qualitative_supp} Right, 
thanks to the construction of full-resolution matching, 
our results best capture local face details and global styles, \textit{e.g.}, face details in the 1st, 3rd, and 5th rows, hand details in the 2nd row, beards in the 4th, 7th, and 9th rows, hair in the 8th row, mouth color in the 10th row, and global color patterns in the 6th row. 

\begin{figure*}[t]
  \centering
  \includegraphics[width=\textwidth]{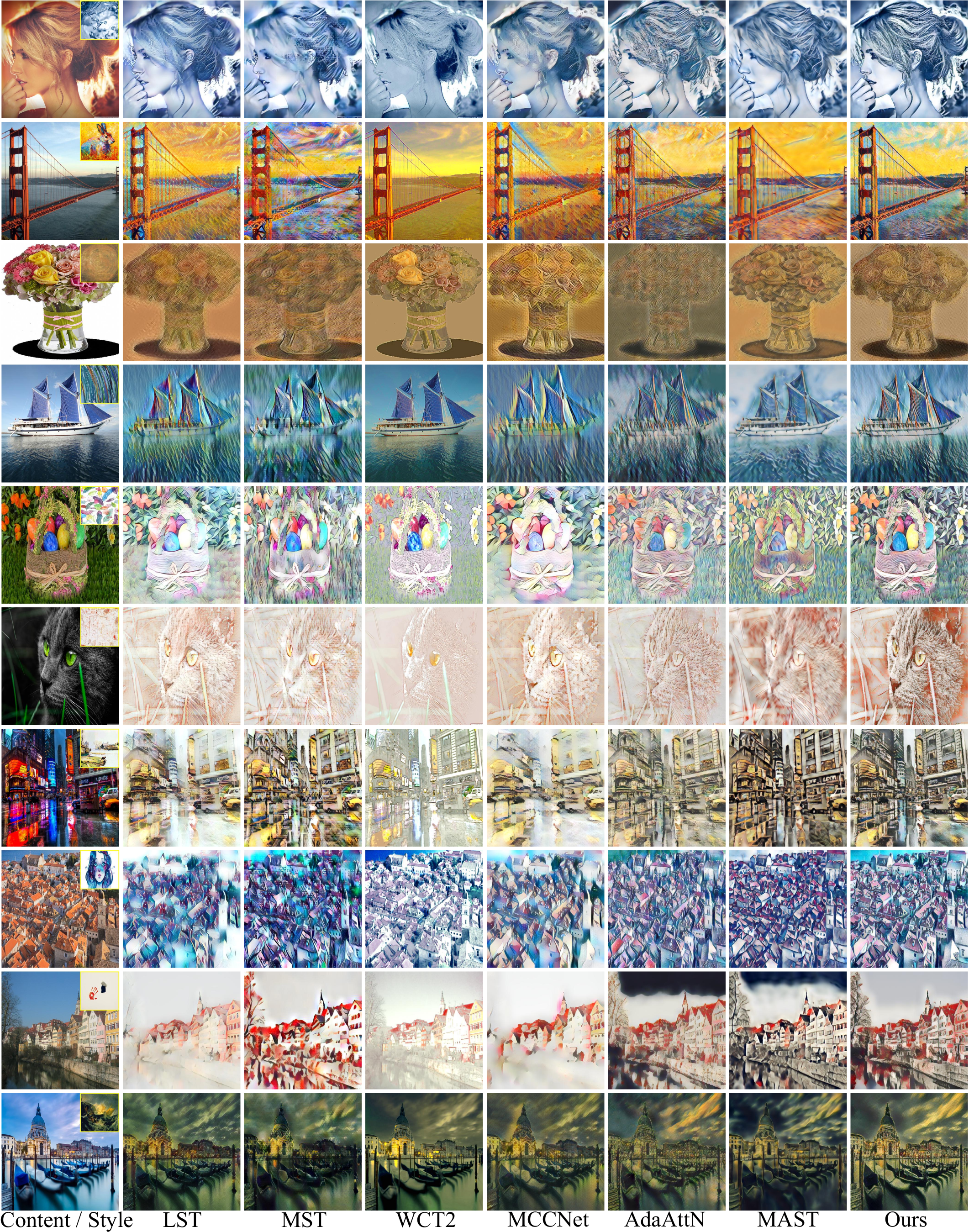}
%   \vspace{-0.7cm}
  \caption{More comparisons with state-of-the-art undistorted style transfer methods.}
  \label{fig:compare_style_supp}
\end{figure*}

\noindent\textbf{Undistorted Image Style Transfer.} 
In Fig.~\ref{fig:compare_style_supp},
we show more comparisons 
with state-of-the-art undistorted style transfer techniques, 
including LST~\cite{li2019learning}, MST~\cite{zhang2019multimodal}, 
WCT2~\cite{yoo2019photorealistic}, MCCNet~\cite{deng2020arbitrary}, 
AdaAttN~\cite{liu2021adaattn}, and MAST~\cite{huo2021manifold}. 
It turns out that the full-resolution matching mechanism in DynaST significantly improves 
the preservation of local details, 
such as the 1st, 3rd, 5th, and 6th rows, 
which in turn enables 
DynaST to better handle more complicated scenes like the 7th, 8th, and 9th rows. 
It also performs well when there 
are extreme textures in the style images, as shown in the 4th row. 
Meanwhile, the migration of global styles also outperforms 
those from previous approaches, \textit{e.g.}, the 2nd and 10th rows. 

\begin{figure*}[t]
  \centering
  \includegraphics[width=\textwidth]{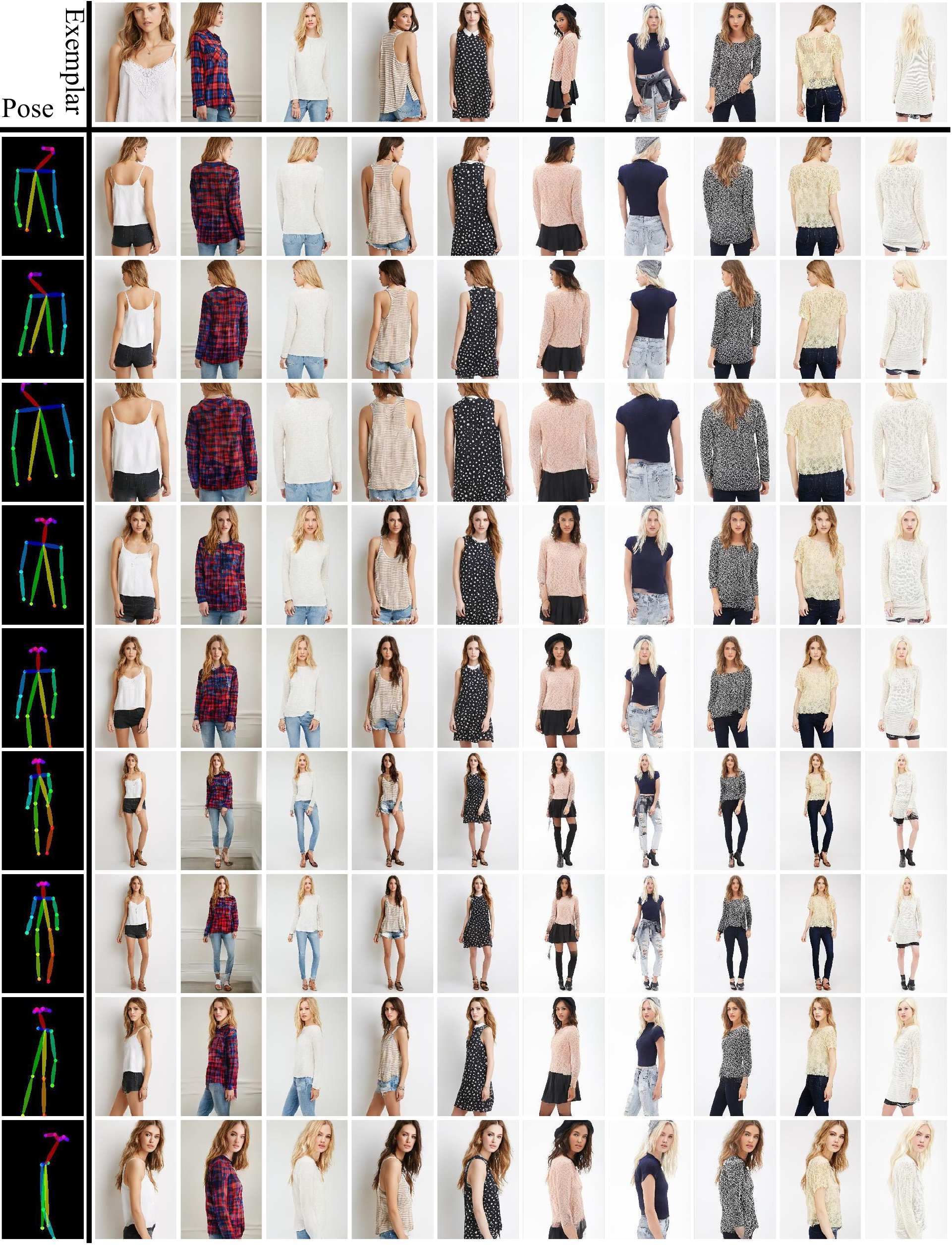}
%   \vspace{-0.7cm}
  \caption{More results by DynaST on pose-guided person image generation.}
  \label{fig:demo_deepfashion_supp}
\end{figure*}

\begin{figure*}[t]
  \centering
  \includegraphics[width=\textwidth]{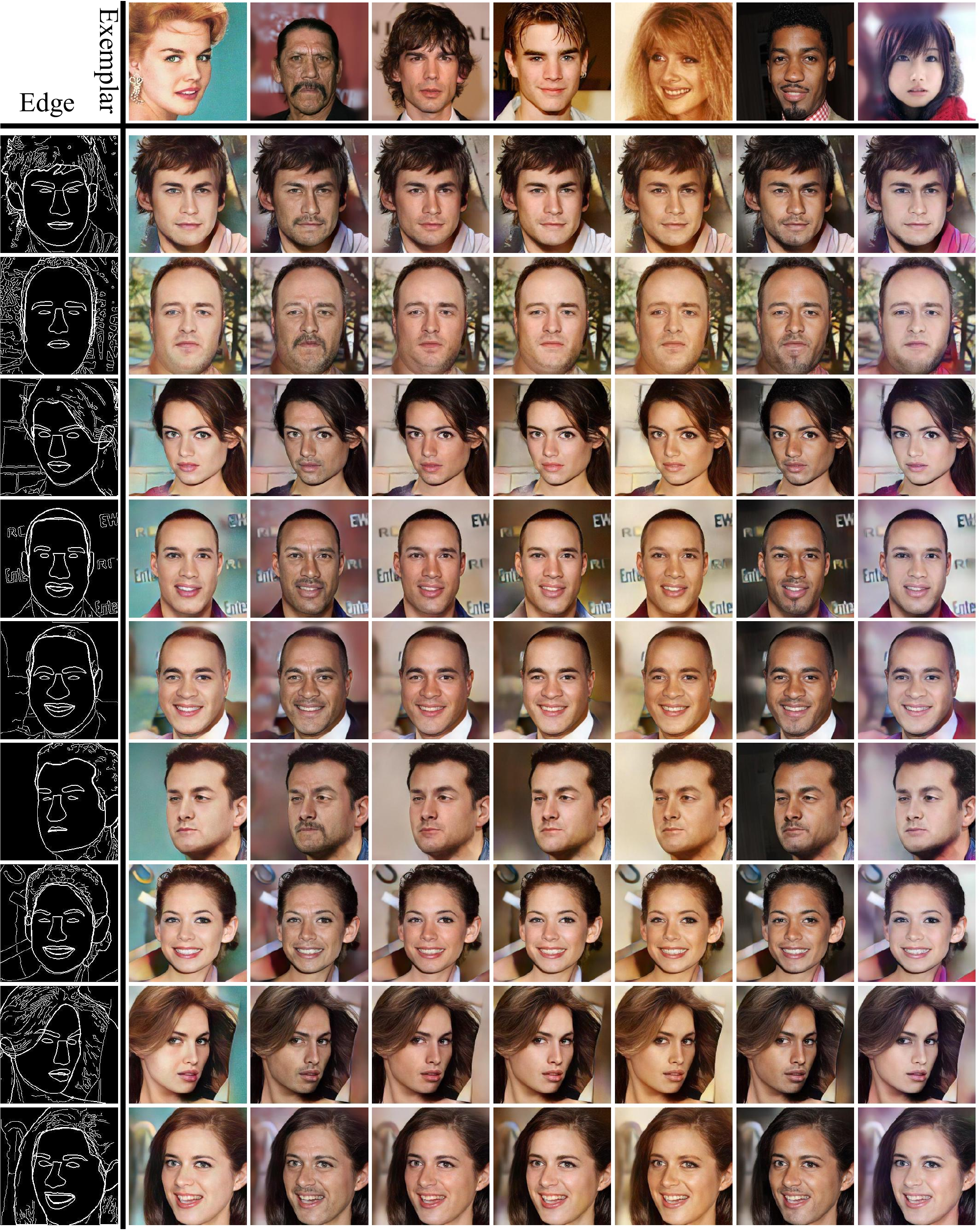}
%   \vspace{-0.7cm}
  \caption{More results by DynaST on edge-based face synthesis.}
  \label{fig:demo_edge2face_supp}
\end{figure*}

\begin{figure*}[t]
  \centering
  \includegraphics[width=\textwidth]{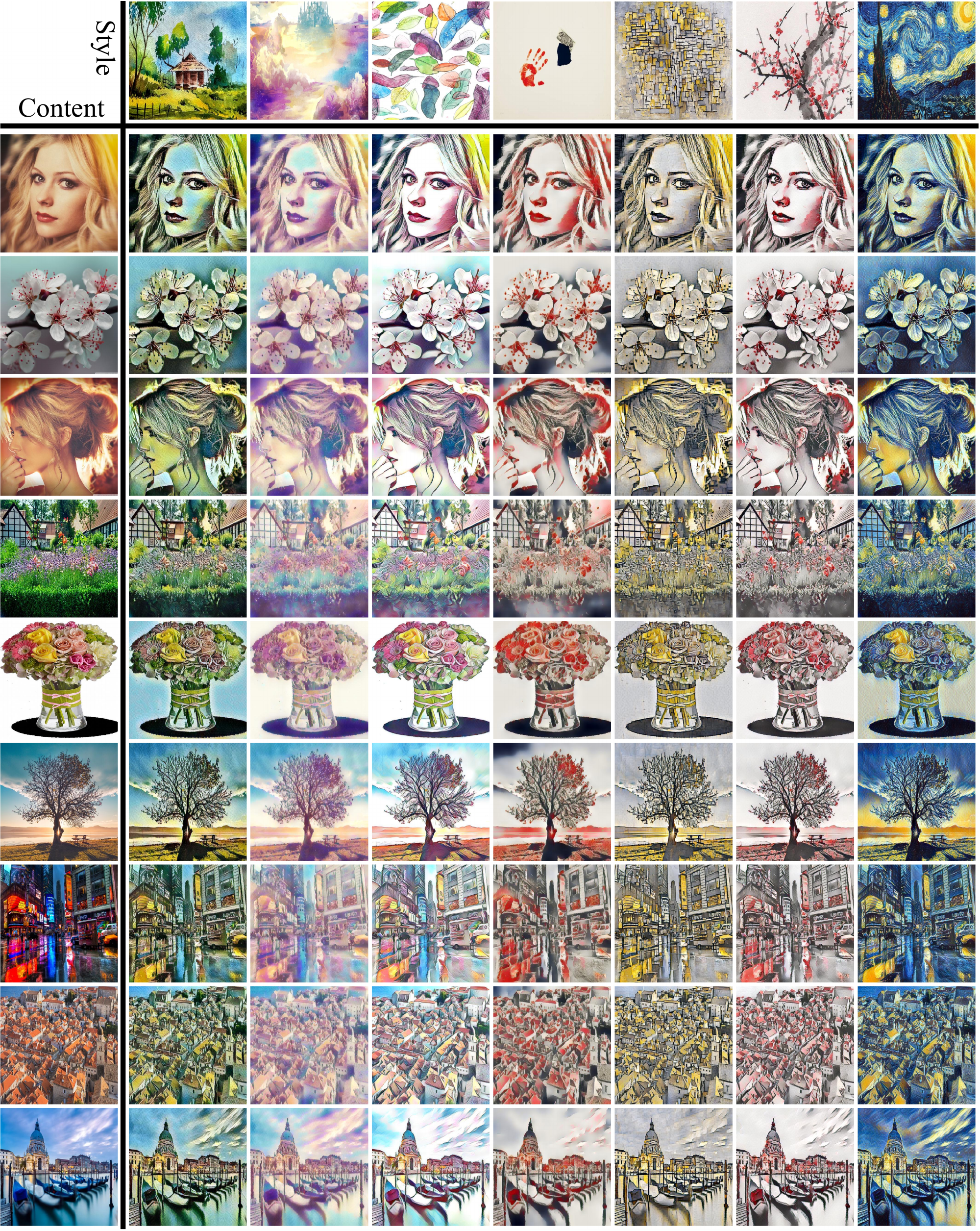}
%   \vspace{-0.7cm}
  \caption{More results by DynaST on undistorted image style transfer.}
  \label{fig:demo_style_supp}
\end{figure*}

\noindent\textbf{Target-Exemplar Pairs.} To further 
illustrate the performance of our proposed DynaST,
we show results under pairwise input semantics and 
exemplar images in Fig. \ref{fig:demo_deepfashion_supp}, 
Fig. \ref{fig:demo_edge2face_supp}, and Fig. \ref{fig:demo_style_supp}
for the three tasks respectively. 
The images are randomly selected, 
which demonstrates the robustness of DynaST to 
different types of input and 
different scale variances between input and exemplar images. 

\end{document}